%% file: main.tex
\documentclass[10pt,twocolumn,letterpaper]{article}

\usepackage{cvpr}              

\definecolor{cvprblue}{rgb}{0.21,0.49,0.74}
\usepackage[pagebackref,breaklinks,colorlinks=red, allcolors=cvprblue]{hyperref}
\usepackage[T1]{fontenc}  
\def\paperID{7282} 
\def\confName{CVPR}
\def\confYear{2025}

\title{Teller: Real-Time Streaming Audio-Driven Portrait Animation with Autoregressive Motion Generation}

\author{Dingcheng Zhen \quad Shunshun Yin \quad Shiyang Qin \quad Hou Yi\\
Ziwei Zhang \quad Siyuan Liu \quad Gan Qi  \quad Ming Tao\\
Shanghai Soulgate Techonolgy Co.tl.\\
{\tt\small {\{dingchengzhen, yinshunshun, qinshiyang, houyi,zhangziwei, siyuanliu, ganqi, ming\}}@soulapp.cn}
}

\begin{document}
\twocolumn[{
\renewcommand\twocolumn[1][]{#1}%
\maketitle
\begin{center}
\centering
\vspace{-20pt}
  \includegraphics[width=0.95\textwidth]{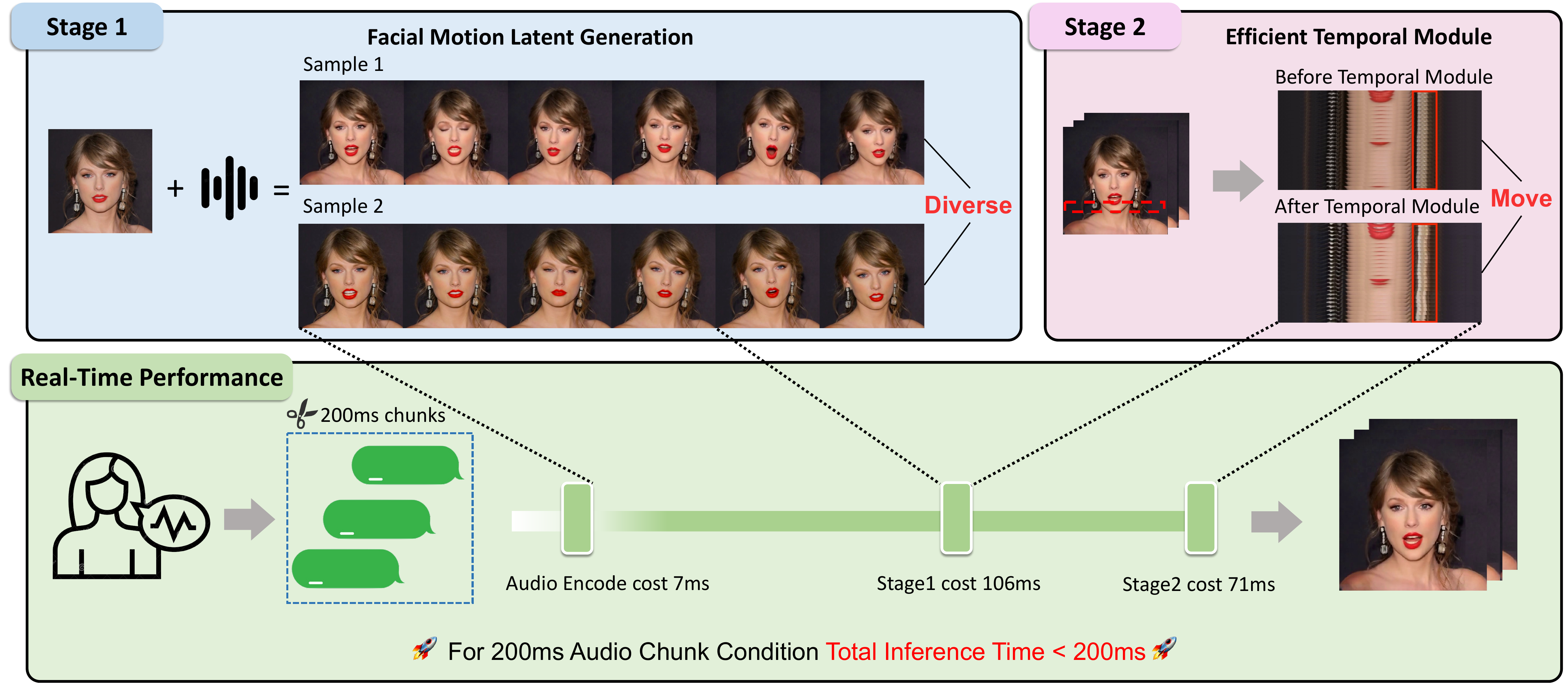}
\captionsetup{font=small}
\end{center}
\vspace{-16pt}
\captionof{figure}{\textbf{Teller} framework is the first autoregressive framework for \textbf{real-time}, audio-driven portrait animation, achieving up to \textbf{25 FPS} while preserving realistic body part and accessory movements. Demo can be found at \url{https://teller-avatar.github.io/}.}
\vspace{4pt}
\label{fig:Teaser_figure}
}]

\input{sec/0_abstract}    
\input{sec/1_intro}
\begin{figure*}[t!]
    \centering
    \includegraphics[width=0.99\textwidth]{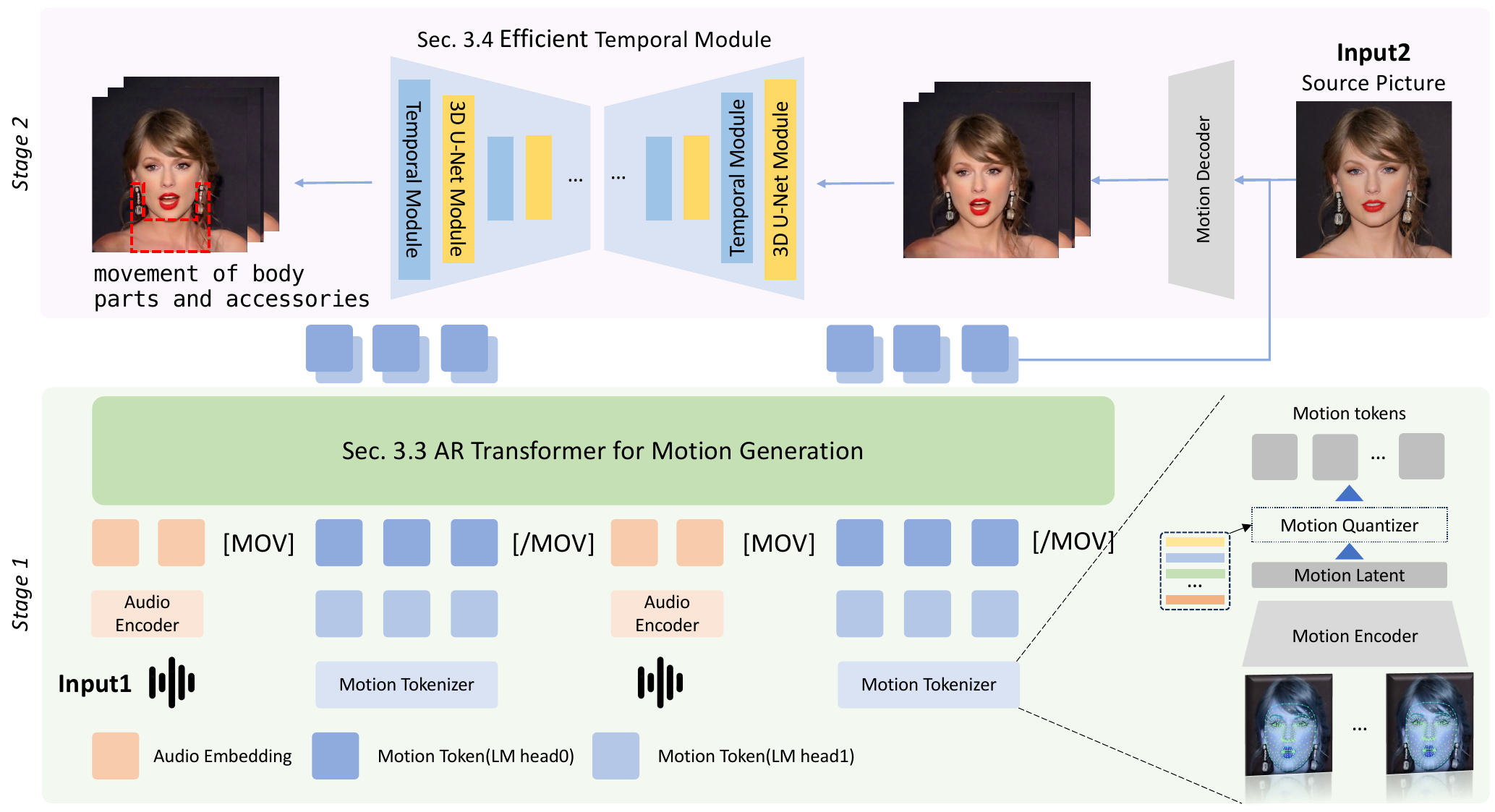}
    \vspace{-5pt}
    \caption{Overall framework of our proposed Teller for real-time streaming audio-driven portrait animation.}
    \vspace{-12pt}
    \label{fig:overall}
\end{figure*}

\input{sec/2_related}

\input{sec/3_method}

\input{sec/4_exp}
\input{sec/5_ab}

\section{Conclusion}
In this paper, we presented \textbf{Teller}, the first autoregressive framework designed for real-time, audio-driven portrait animation. Addressing the challenge of realistic and efficient talking head generation, Teller achieves high-quality animations at up to 25 FPS, surpassing existing methods in both fidelity and responsiveness. 
Extensive experiments demonstrated Teller’s advantages over SoTA audio-driven animation methods, particularly in rendering nuanced movements essential for lifelike and visually convincing animations. Human evaluations further validate its quality, particularly in natural expression and lip synchronization.  By balancing computational efficiency with high animation fidelity, Teller sets a new standard for real-time talking head animation, marking a significant advancement in multimodal portrait animation frameworks.
Additionally, Teller’s AR Transformer architecture makes it compatible with existing unified multimodal language models.
\clearpage
{
    \small
    \bibliographystyle{ieeenat_fullname}
    \bibliography{main}
}
\end{document}

%% file: sec/0_abstract.tex
\begin{abstract}
In this work, we introduce the \textbf{first} autoregressive framework for real-time, audio-driven portrait animation, \textit{a.k.a}, talking head. Beyond the challenge of lengthy animation times, a critical challenge in realistic talking head generation lies in preserving the natural movement of diverse body parts. 
To this end, we propose \textbf{Teller}, the first streaming audio-driven protrait animation framework with autoregressive motion generation. 
Specifically, \textbf{Teller} first decomposes facial and body detail animation into two components: Facial Motion Latent Generation (FMLG) based on an autoregressive transfromer, and movement authenticity refinement using a Efficient Temporal Module (ETM).
Concretely, FMLG employs a Residual VQ model to map the facial motion latent from the implicit keypoint-based model into discrete motion tokens, which are then temporally sliced with audio embeddings. This enables the AR tranformer to learn real-time, stream-based mappings from audio to motion.
Furthermore, \textbf{Teller} incorporate ETM to capture finer motion details. This module ensures the physical consistency of body parts and accessories, such as \textbf{neck muscles and earrings}, improving the realism of these movements.
\textbf{Teller} is designed to be efficient, surpassing the inference speed of diffusion-based models (Hallo \textbf{20.93s} vs. \textcolor{red}{Teller \textbf{0.92s}} for one second video generation), and achieves a real-time streaming performance of up to \textbf{\textcolor{red}{25 FPS}}. Extensive experiments demonstrate that our method outperforms recent audio-driven portrait animation models, especially in small movements, as validated by human evaluations with a significant margin in quality and realism.
\end{abstract}

%% file: sec/1_intro.tex
\vspace{-6pt}
\section{Introduction}
\label{sec:intro}
Realistic and expressive portrait animation from audio and static images, commonly known as talking head animation~\cite{hu2024animate, wang2021one, guo2024liveportrait, lu2021live}, has garnered significant interest across applications such as virtual avatars, digital communication, and entertainment. However, generating high-quality animations that are visually compelling and temporally consistent remains a major challenge. This complexity stems from the need to intricately coordinate lip movements, facial expressions, and head positioning to create lifelike effects. Moreover, achieving real-time, realistic talking head animation is complicated by computational constraints and the nuances of human movement, making this an especially demanding task.

While recent advancements, such as diffusion models~\cite{shen2023difftalk, yao2024fd2talk, stypulkowski2024diffused, sun2024diffposetalk}, have improved high-quality content generation, achieving controllable animation presents ongoing hurdles. Effective animation requires the accurate capture of complex facial expressions, body gestures, and the subtle interplay between them. Existing methods often suffer from prolonged animation times (See animation time in Tab~\ref{tab:HDTF}), limiting their potential for real-time applications, and frequently fail to capture the natural, interconnected motions of various facial and body parts, \eg, earrings and necklaces, as shown in Figure.~\ref{fig:Teaser_figure} stage 2. This shortfall results in animations with stiff or exaggerated movements that disrupt the realism of the animation (See Figure~\ref{fig:movements}). Addressing these challenges necessitates a solution that balances computational efficiency with high animation quality without overloading processing resources.

To this end, we propose \textbf{Teller}, the \textbf{first} autoregressive framework capable of real-time, streaming-based talking head animation at up to 25 FPS. As shown in Figure.~\ref{fig:overall}, Teller employs a two-stage framework, combining \textbf{F}acial \textbf{M}otion \textbf{L}atent \textbf{G}eneration (\textbf{FMLG}) and an \textbf{E}fficient \textbf{T}emporal \textbf{M}odule (\textbf{ETM}) to produce realistic and physically consistent animations across facial and body movements. In the first stage, FMLG uses a Residual Vector Quantization (RVQ) model~\cite{zeghidour2021soundstream} to encode facial motion latents derived from an implicit keypoint-based model into discrete motion tokens. These tokens are then temporally aligned with audio embeddings, allowing an autoregressive (AR) transformer to map audio signals to facial movements in real-time. By breaking down the motion into temporal segments, FMLG enables the AR transformer to dynamically and efficiently generate high-quality animations responsive to live audio inputs.

To enhance the realism of body movements, Teller introduces ETM in the second stage, which captures subtle details that are often overlooked in existing methods~\cite{guo2024liveportrait, xu2024vasa}. The ETM refines finer movements, such as body parts and accessories, ensuring physically plausible interactions. For instance, it simulates realistic motions in neck muscles and dynamic accessories like earrings, which are vital for maintaining visual continuity in the animated avatar.
Our Teller model prioritizes computational efficiency and is specifically designed to significantly outperform diffusion-based models in terms of inference speed. For example, generating a one-second video takes Hallo \textbf{20.93s}, while Teller completes the task in just \textbf{\textcolor{red}{0.92s}} (see Tab.~\ref{tab:HDTF}). Despite this speed, Teller maintains high animation fidelity. By optimizing computational demands without compromising output quality, Teller not only meets but exceeds real-time requirements, ensuring a smooth and responsive user experience. 

In extensive evaluations across various benchmarks and real-world settings, Teller demonstrates significant improvements over current state-of-the-art audio-driven portrait animation models, as shown in Table.~\ref{tab:HDTF}, particularly in capturing nuanced facial and body movements. Human evaluations validate Teller's superior quality and realism, especially in rendering the subtle movements essential for lifelike animations. Our research represents a notable advancement in real-time talking head animation, presenting an innovative framework that bridges the gap between realism and efficiency in multimodal animation, as depicted in Figure.~\ref{fig:sample_diversity}, Figure.~\ref{fig:movements}, and Figure.~\ref{fig:Lip_acc}.


%% file: sec/2_related.tex
\section{Related Works}
\label{sec:2}
\noindent \textbf{Non-Diffusion-Based Audio-Driven Portrait Animation}
This line of works typically consist of two key components: an audio-to-motion model and a facial motion representation model. These methods often utilize implicit keypoints as an intermediate motion representation, warping the source portrait based on the driving image. The goal of implicit methods is to learn disentangled representations in 2D~\cite{burkov2020neural,liang2022expressive,pang2023dpe,wang2023progressive,yin2022styleheat,zhou2021pose}or 3D ~\cite{drobyshev2022megaportraits,wang2021one}latent spaces, with a focus on aspects such as identity, facial dynamics, and head pose. For example, FOMM~\cite{siarohin2019first} employs first-order Taylor expansion to capture local motion, while FaceVid2Vid~\cite{wang2021one} extends this by introducing a 3D implicit keypoint representation, enabling free-view portrait animation.

To effectively learn facial motion representations in latent space, these approaches often rely on GAN-based frameworks to disentangle identity-related appearance from non-identity-related motion, specifically capturing expressions, lip and eye movements, minor accessories, and poses. Examples include FaceVid2Vid~\cite{wang2021one}, LivePortrait~\cite{guo2024liveportrait}, and others, where identity and motion representations are independently learned to generate realistic talking head animations.
For instance, MakeItTalk~\cite{zhou2020makelttalk} uses an LSTM-based audio-to-motion model to predict landmark coordinates from audio input, which are then translated into video frames using a warp-based GAN model. Similarly, SadTalker~\cite{Sadtalker} employs FaceVid2Vid~\cite{wang2021one} as an image synthesizer, with ExpNet and PoseVAE modules transforming audio features into inputs compatible with FaceVid2Vid for audio-to-video generation.

However, these approaches face challenges due to GAN losses that primarily focus on facial expressions, lip, and eye movements, often neglecting accessory, hair, and body movements. This can result in animations that appear stiff or incomplete, lacking natural dynamism.
In contrast, our work introduces the \textbf{first} autoregressive framework specifically designed for real-time, audio-driven portrait animation, achieving up to 25 FPS and delivering a more realistic, coherent portrayal of both facial and accessory movements.

\noindent \textbf{Diffusion-based Audio-driven Portrait Animation}
Recent advancements in diffusion-based video generation have demonstrated promising outcomes for audio-driven portrait animation. Methods like GAIA~\cite{he2023gaia} and VASA-1~\cite{xu2024vasa} have designed diffusion models to transform audio inputs into motion latents, facilitating audio-to-video generation. Further developments, such as EMO\cite{tian2024emo}, Hallo~\cite{xu2024hallo}, LOOPY~\cite{jiang2024loopy}, enhance end-to-end diffusion modeling by incorporating motion modules~\cite{blattmann2023align, guo2023animatediff,  hu2024animate} and audio cross-attention mechanisms, improving the coherence and synchronization between audio cues and visual motion.
Despite these improvements, a significant limitation of diffusion-based models remains their multi-step inference process, required to generate even a single frame or a few frames of video. This step-by-step prediction approach renders real-time performance challenging, as it is computationally intensive and time-consuming, making diffusion-based models less suitable for applications demanding instantaneous response.

\noindent \textbf{AR Transformer-Based Generation}
Recent works~\cite{wang2024emu3, zhou2024transfusion, xie2024show, tang2024any, ye2024x, aiello2023jointly} have focused on developing unified multimodal language models for generating visual content, such as images and videos. Some studies~\cite{zhu2023vl, sun2023generative} use autoregressive modeling with continuous representations interleaved with text tokens for image generation. Others, like SEED-X~\cite{ge2024seed}, propose a foundational system combining CLIP ViT-based image representations with text tokens for multimodal tasks, including next-token prediction and image representation regression. DreamLLM~\cite{dong2023dreamllm} also explores multimodal understanding and creation, while Chameleon~\cite{team2405chameleon} introduces token-based models capable of both understanding and generating images.

%% file: sec/3_method.tex
\section{Method}

As shown in Fig.~\ref{fig:overall}, Teller comprises two main modules: the Facial Motion Latent Generation (\textbf{FMLG}) and the Efficient Temporal Module (\textbf{ETM}). FMLG module integrates an autoregressive transformer and a residual vector quantization (RVQ) component. It applies an autoregressive transformer to generate discrete facial motion tokens from audio input. Following FMLG, ETM refines the generated motion to produce realistic body and accessory movements, ensuring physical consistency in animated results.

\subsection{Preliminaries}


    

Prior works, such as LivePortrait~\cite{guo2024liveportrait}  have introduced methods for extracting implicit keypoints as facial motion latents using motion and appearance extractors. These motion latents capture essential facial dynamics needed for animating input images and consist of three main components:
\begin{itemize}
    \item \textbf{Expression Deformation} ($\delta$): A set deformation of 21 implicit keypoints, represented as \( {\delta} = [{\delta}_1, {\delta}_2, \dots, {\delta}_{21}] \), where each \( {\delta}\) is a 3D vector (\( \mathbb{R}^3 \)) that indicates the position of the \( i \)-th deformation of  facial keypoint.
    
    \item \textbf{Head Pose} (\( R \)): Defined by three rotation vectors \( R = [r_1, r_2, r_3] \), with each \( r_i \) being a 3D vector (\( \mathbb{R}^3 \)) that describes the head's orientation in 3D space.

    \item \textbf{Expression Deformation} (\( t \)): A single 3D vector (\( t \in \mathbb{R}^3 \)) that captures facial expression deformations.
\end{itemize}
These components are concatenated into a unified motion latent of size \( 25 \times 3 \):
\begin{equation}
\setlength{\abovedisplayskip}{3pt}
\setlength{\belowdisplayskip}{3pt}
 m = [{\delta}_1,{\delta}_2, \dots, {\delta}_{21}, r_1, r_2, r_3, t],
\end{equation}
where \( m \in \mathbb{R}^{25 \times 3} \) includes the 21 keypoints, head pose, and expression deformation.

\subsection{Facial Motion Latent Generation (FMLG)}
In FMLG, Teller generates motion latent from facial motion extraction, then encodes it into discrete tokens using a residual vector quantizer (RVQ). To optimize encoding, the concatenated motion latent $m$ is processed with RVQ, leveraging temporal redundancy across $T$ frames for efficient compression.
The RVQ quantization loss, which encodes the continuous $m$ into discrete tokens, is defined as:
\begin{equation}
\setlength{\abovedisplayskip}{3pt}
\setlength{\belowdisplayskip}{3pt}
\small
  \mathcal{L}_{vq} = \sum_{t=1}^{T} \left[
\underbrace{||m - \text{FFN}_{dec}(z_t + \text{sg}[\hat{z_t} - z_t])||_2^2}_{\mathcal{L}_{\text{recon}}} + 
\underbrace{||z_t - \text{sg}[\hat{z_t}]||_2^2}_{\mathcal{L}_{\text{commit}}} 
\right]  
\end{equation}
where $z_t = \text{FFN}_{enc}(m)$ represents the encoded latent at each time step $t$, $\hat{z_t}$ is the quantized latent, and $\text{sg}$ denotes the stop-gradient operation. The two loss components, $\mathcal{L}_{\text{recon}}$ and $\mathcal{L}_{\text{commit}}$, are defined as follows:

\begin{itemize}
    \item \textbf{Reconstruction Loss} ($\mathcal{L}_{\text{recon}}$): Minimizes the difference between the original motion latent $m$ and the decoded quantized latent.
    \item \textbf{Commitment Loss} ($\mathcal{L}_{\text{commit}}$): Encourages $z_t$ to approach the quantized latent $\hat{z_t}$, ensuring stability in quantization.
\end{itemize}

This approach enables FMLG to learn robust, temporally-consistent representations, translating audio-driven inputs into lifelike facial animations. Experimentally, to achieve trade-offs between frame count and redundancy, we selected 4 frames (4$\times$25$\times$3 latent)  to be compressed into 32 tokens (hyper-parameter selection trade-off refer to Fig.~\ref{fig:VQ_ablation}).

\begin{figure}
    \centering
    \includegraphics[width=\linewidth]{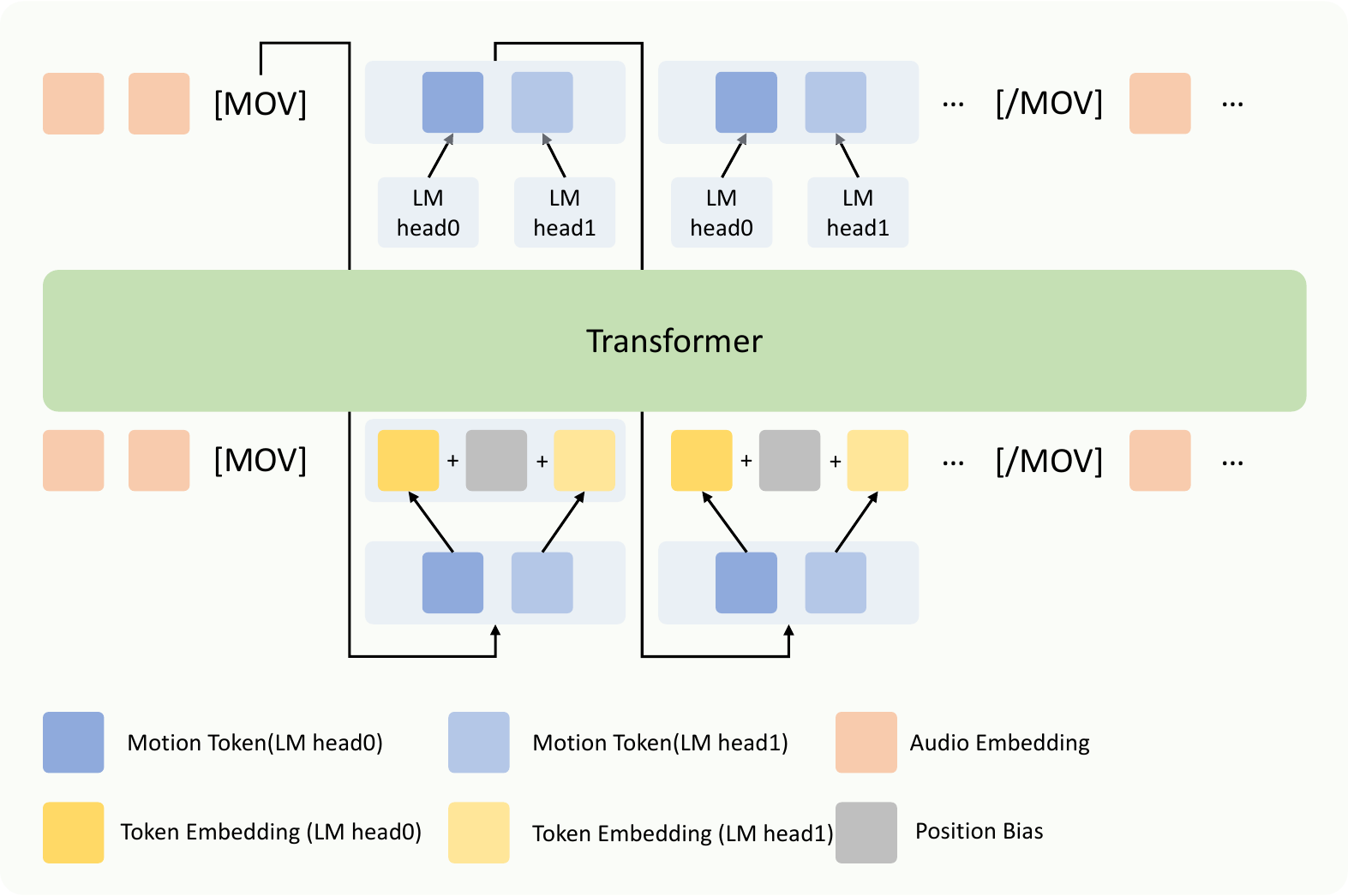}
    \caption{In our Teller, we follow AR transformer architecture, but each input consists of a pair of tokens and model pred a pair of tokens for each output position.}
    \label{fig:AR_transformer}
    \vspace{-16pt}
\end{figure}

\begin{figure*}[th!]
    \centering
    \includegraphics[width=\textwidth]{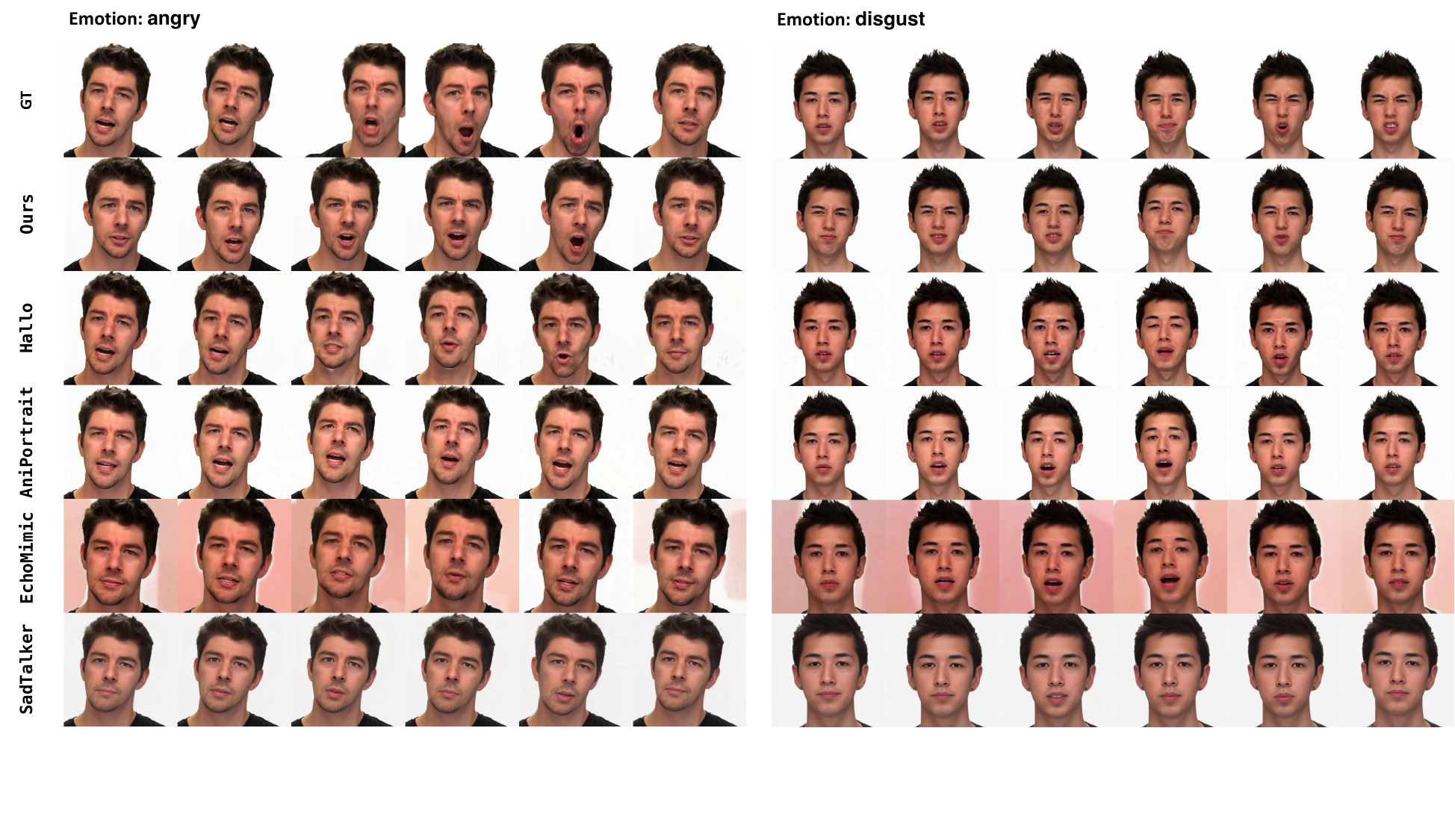}
    \vspace{-55pt}
    \caption {Qualitative comparison with existing approaches on RAVDESS data-set of 'angry' and 'disgust' emotion cases. \textit{\textbf{\textcolor{red}{Videos} are available in the supplementary materials.}}}
        \label{fig:Qualitative_RAVDESS_angry}
    \vspace{-14pt}
\end{figure*}

\subsection{AR Transformer for Motion Generation}

Using the learned RVQ-based latents, motion is denoted as:
\begin{equation}
\setlength{\abovedisplayskip}{3pt}
\setlength{\belowdisplayskip}{3pt}
    M = [m_1, m_2, \dots, m_T], \quad \text{where} \quad m_i \in \mathbb{R}^{25 \times 3}.
\end{equation}
This sequence of motion latents is converted to discrete tokens using the RVQ module:
\begin{equation}
\setlength{\abovedisplayskip}{3pt}
\setlength{\belowdisplayskip}{3pt}
    T_m = [t_1, t_2, \dots, t_{T/4}],
\end{equation}
where $T_m$ represents the quantized motion tokens. 
Audio input is encoded with the Whisper encoder~\cite{cao2012whisper}, generating the audio condition \( c \). Motion generation is modeled as a next-token prediction task, where the distribution of each token is predicted based on the previous \( t-1 \) tokens and audio condition \( c \):
\begin{equation}
\setlength{\abovedisplayskip}{3pt}
\setlength{\belowdisplayskip}{3pt}
    P(t_i \mid c, t_{<i}).
\end{equation}
This autoregressive setup enables sequential generation of motion based on past motion tokens and audio embeddings.

To enable real-time streaming animation, we process both audio and video frames in 200ms chunks, following Whisper’s constraints. Each audio chunk is encoded into a \( [10 \times 512] \) embedding, while each video chunk uses 32 learned RVQ-based motion tokens.
For efficient real-time performance, we enable the autoregressive transformer to process token pairs at each position, which improves prediction speed by processing two tokens concurrently. In Teller, each input consists of a token pair with a combined embedding and a learnable position bias inspired by BERT, capturing relative token positions.
The loss for each head, representing each token in the pair, is computed as:
\begin{equation}
\setlength{\abovedisplayskip}{3pt}
\setlength{\belowdisplayskip}{3pt}
\mathcal{L}_{\text{head0}_{j}} = \text{CE}(\text{label}_{j}[0], \text{T}(\text{input}_{j}[0] | \text{input}_{<j})),
\end{equation}
\begin{equation}
\setlength{\abovedisplayskip}{3pt}
\setlength{\belowdisplayskip}{3pt}
\mathcal{L}_{\text{head1}_{j}} = \text{CE}(\text{label}_{j}[1], \text{T}(\text{input}_{j}[1] | \text{input}_{<j})),
\end{equation}
where \( T \) denotes the transformer. The total loss, with a regularization term to balance learning across both heads, is:
\begin{equation}
\setlength{\abovedisplayskip}{3pt}
\setlength{\belowdisplayskip}{3pt}
\mathcal{L}_{ar} = \sum_{j=1}^{I/2} \left[ \mathcal{L}_{\text{head0}_{j}} + \mathcal{L}_{\text{head1}_{j}} + \left\|\mathcal{L}_{\text{head0}_{j}} - \mathcal{L}_{\text{head1}_{j}}\right\|_2^2 \right].
\end{equation}
This regularization term, \( \left\|\mathcal{L}_{\text{head0}} - \mathcal{L}_{\text{head1}}\right\|_2^2 \), ensures balanced training across the two heads, promoting stable and accurate real-time animation.

After testing frame count and redundancy, we use 4 frames (4$\times$25$\times$3 latent), compressed into 32 tokens (refer to Fig.~\ref{fig:VQ_ablation}). Then the 4 frames are interpolated to 5 frames for faster generation.
This setup balances inference efficiency and feature quality, enabling real-time, high-fidelity streaming portrait animation. 

\subsection{Efficient Temporal Module for Refinement}

In diffusion-based models, temporal layers are added to text-to-image (T2I) frameworks to capture frame dependencies~\cite{hu2024animate}. Inspired by this, Teller incorporates temporal refinement but achieves it in a single step, unlike diffusion models that require multiple iterations, enhancing real-time efficiency.
After encoding video frames with a VAE encoder, we apply a 3D U-Net~\cite{cciccek20163d} to extract features from the image sequence, represented as
$x \in \mathbb{R}^{b \times t \times h \times w \times c}$,
where \( b \) is the batch size, \( t \) the frame count, \( h \) and \( w \) the frame dimensions, and \( c \) the channel count. The features are reshaped to
$x \in \mathbb{R}^{(b \times h \times w) \times t \times c}$,
enabling ETM to perform self-attention along the temporal dimension \( t \). ETM's output is then merged with the original features through residual connections, integrating temporal dependencies into spatial features.

For training, we use the first 5 frames from a real image sequence and the subsequent 5 frames reconstructed with LivePortrait~\cite{guo2024liveportrait}. After processing through ETM, we compute reconstruction loss between the predicted and ground-truth frames:
\begin{equation}
\setlength{\abovedisplayskip}{3pt}
\setlength{\belowdisplayskip}{3pt}
\mathcal{L}_{\text{recon}} = \sum_{i=6}^{10} \left\| x_{\text{gt}_i} - f(x_i | x_{\text{gt}_{<6}}) \right\|_2^2,
\end{equation}
where \( x_{\text{gt}_i} \) denotes the real feature sequence and \( x_i \) the reconstructed feature sequence (from LivePortrait or the Stage 1 decoder).
ETM primarily learns to preserve consistency in physical features such as neck muscles and earrings. To this end, we add a region-specific mask to reconstruction loss, and the final loss of ETM:
\begin{equation}
\mathcal{L}_{\text{ETM}} = \sum_{i=6}^{10} \left\| x_{\text{gt}_i} \odot \text{mask}_i - f(x_i | x_{\text{gt}_{<6}}) \odot \text{mask}_i \right\|_2^2,
\end{equation}
where element-wise multiplication with \(\text{mask}_i\) focuses reconstruction on specific regions. The mask is defined as:
\begin{equation}
\setlength{\abovedisplayskip}{3pt}
\setlength{\belowdisplayskip}{3pt}
\text{mask}(i, j) = 
  \begin{cases}
    1, & \text{if } (i, j) \text{ is within BB}(x) \\
    0, & \text{otherwise}
  \end{cases}
\end{equation}
where BB(x) stands for the bounding box and outlines relevant body parts. Key landmarks are identified using MediaPipe\cite{lugaresi2019mediapipe}, with points indices such as [93, 323, 152] defining the bounding boxes for these regions, highlighted in red in the stage 2 input of Fig.~\ref{fig:overall}.

%% file: sec/4_exp.tex
\begin{figure}[t!]
    \centering
    \includegraphics[width=\linewidth]{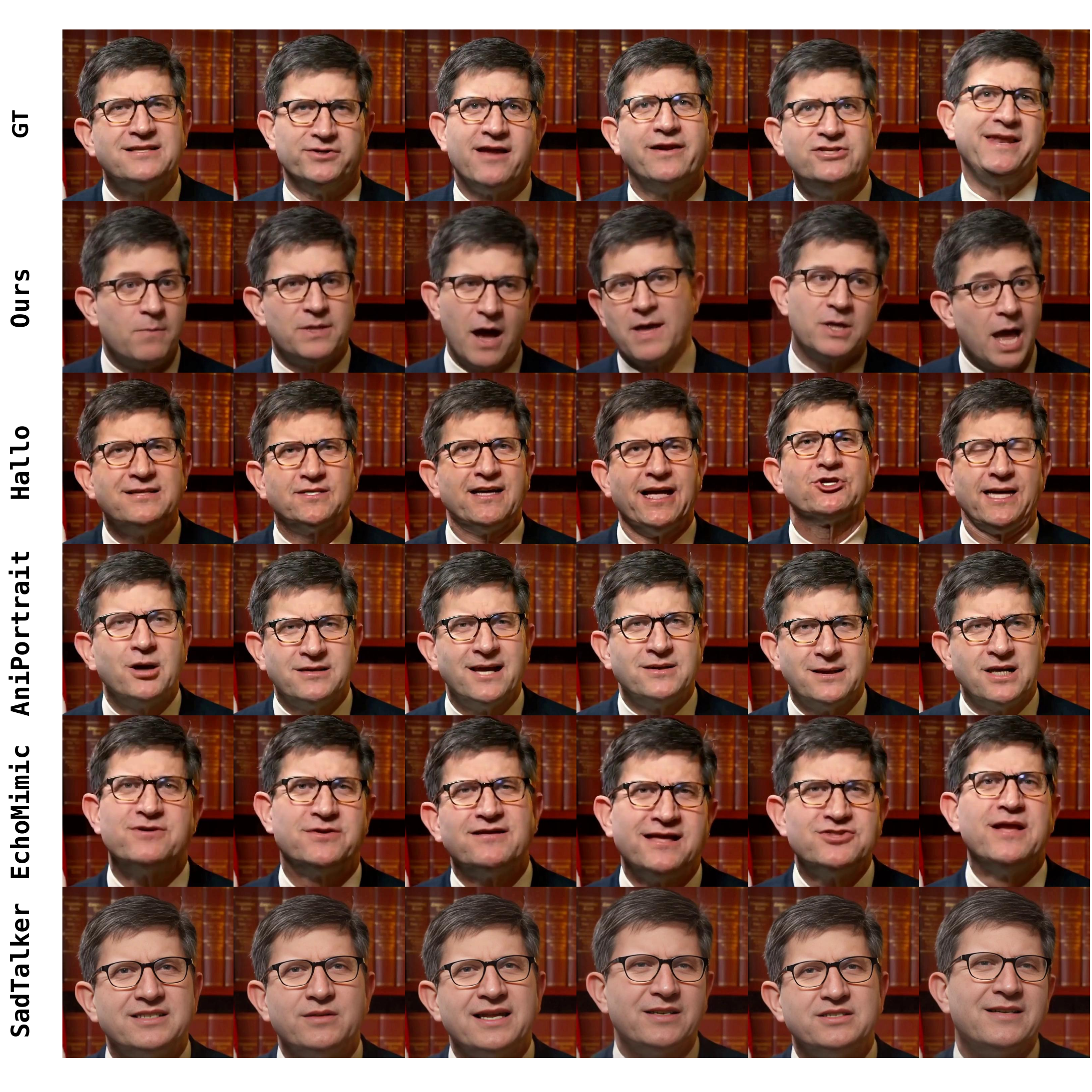}
    \vspace{-24pt}
    \caption {Qualitative comparison with existing approaches on HDTF dataset. \textit{\textbf{\textcolor{red}{Videos} are available in the supplement. mat.}}}
        \label{fig:Qualitative_HDTF_Head}
    \vspace{-16pt}
\end{figure}

\section{Experiments}
\subsection{Experimental Setup}
\noindent \textbf{Datasets.} For training, we used the AV Speech datasets (filtered to 662 hours)~\cite{Ephrat_2018} and VFHQ~\cite{xie2022vfhq} datasets (filtered to 2 hours) for pretraining, along with additional talking-head videos from the internet (32 hours) for supervised fine-tuning (SFT). For validation, we used the HDTF (filtered to 0.83 hours)~\cite{zhang2021flow} and RAVDESS (filtered to 0.55 hours)~\cite{livingstone2018ryerson} datasets and supplementary internet data (0.49 hours) for qualitative comparisons and human evaluation only.
To ensure quality, we applied the Mediapipe~\cite{lugaresi2019mediapipe} face detection tool to filter out instances with facial movement exceeding 50\%. We further refined the data using Sync-C and Sync-D to exclude samples with low lip-sync scores.

\noindent \textbf{Metrics.} Evaluation metrics include Fréchet Inception Distance (FID)\cite{heusel2017gans}, Fréchet Video Distance (FVD)\cite{unterthiner2018towards}, Synchronization-C (Sync-C)\cite{prajwal2020lip}, and Synchronization-D (Sync-D)\cite{prajwal2020lip}. FID and FVD assess realism, with lower scores indicating better quality, while Sync-C and Sync-D measure lip synchronization, with higher Sync-C and lower Sync-D values indicating better alignment.

\noindent \textbf{Implementation Details.} 

\noindent \textbf{Stage 1}: During pretraining, we follow the architecture design of the Qwen1.5-4B model~\cite{bai2023qwen} and initialize the parameters randomly. The model is trained on an 8$\times$8 Nvidia A800 GPU machine with a batch size of 1024, using the AdamW optimizer~\cite{loshchilov2017decoupled}. We employ a cosine learning rate scheduler, with the learning rate decaying from 1e-4 to 1e-6 over 40 epochs.
In the supervised fine-tuning (SFT) phase, we again use an 8$\times$8 Nvidia A800 GPU machine, but with a batch size of 512. The AdamW optimizer~\cite{loshchilov2017decoupled}  is used, along with a cosine learning rate scheduler. The learning rate decaying from 1e-5 to 1e-6 over 10 epochs.
\noindent \textbf{Stage 2}: The model is trained on an 8$\times$8 Nvidia A800 GPU machine, with a batch size of 1024, and using the AdamW optimizer. The cosine learning rate scheduler is used, with the learning rate decaying from 1e-4 to 1e-6 over 30 epochs.

\noindent \textbf{Real time analysis.} The model is inference on an 4 Nvidia H800 GPU machine. For a 200ms audio input, the average processing time of the Whisper encoder is 7ms. In Stage 1, the average total time is 106ms, with the AR transformer taking an average of 6ms per 16 tokens, and the motion decoder taking an average of 10ms. In Stage 2, the average total time is 71ms, with the VAE encoder and decoder averaging 25ms, and the Temporal Module averaging 21ms. 

\begin{table}[]
\centering
\resizebox{\linewidth}{!}{
\begin{tabular}{lcccc|c}
\toprule
\textbf{Method} & \textbf{FID$\downarrow$} & \textbf{FVD$\downarrow$} & \textbf{Sync-C$\uparrow$} & \textbf{Sync-D$\downarrow$}  & $Time$ \\
\midrule
SadTalker~\cite{Sadtalker}     & 22.177& 233.673& 7.326& 7.848 &18.89s\\
EchoMimic~\cite{chen2024echomimic}     & 23.049& 290.190& 6.664& 8.839 &31.10s\\
AniPortrait~\cite{wei2024aniportrait}   & 28.161& 235.099& 4.547& 10.657 &29.36s\\
Hallo~\cite{xu2024hallo}         & \textbf{20.639}& \underline{174.191}& \underline{7.497}& \underline{7.741} &20.93s\\
Teller (ours)          & \underline{21.352}& \textbf{173.463}& \textbf{7.696}& \textbf{7.536} &\textbf{\textcolor{red}{0.92s}}\\ \midrule
Real video    & -      & -       & 8.094& 6.976 & - \\
\bottomrule
\end{tabular}}
\vspace{-8pt}
\caption{Quantitative comparison with existing portrait image animation approaches on the HDTF dataset. $Time$ stands for the averaging time cost of generating one second of 25 fps video.
}
\label{tab:HDTF}
\vspace{-12pt}
\end{table}

\begin{table}[]
\centering
\resizebox{\linewidth}{!}{
\begin{tabular}{lcccc}
\toprule
\textbf{Method} & \textbf{FID$\downarrow$} & \textbf{FVD$\downarrow$} & \textbf{Sync-C$\uparrow$} & \textbf{Sync-D$\downarrow$} \\
\midrule
SadTalker~\cite{Sadtalker}     & 32.343& 487.924& \underline{4.304}& \textbf{7.621}\\
EchoMimic~\cite{chen2024echomimic}     & 21.058& 668.675& 3.292& 9.096\\
AniPortrait~\cite{wei2024aniportrait}   & 30.696& \underline{476.197}& 2.321& 11.542\\
Hallo~\cite{xu2024hallo}         & \textbf{19.826}& 537.478& 4.062& 8.552\\
Teller (ours)          & \underline{20.352}& \textbf{429.288}& \textbf{4.496}& \underline{7.936}\\ \midrule
Real video    & -      & -       & 5.223& 7.069\\
\bottomrule
\end{tabular}}
\vspace{-8pt}
\caption{Quantitative comparison with existing portrait image animation approaches on the RAVDESS dataset. }
\label{tab:RAVDESS}
\vspace{-12pt}
\end{table}

\subsection{Quantitative Results}

\noindent \textbf{Comparison on the HDTF Dataset.} Table~\ref{tab:HDTF} presents quantitative results for portrait animation techniques on the HDTF dataset. Our method outperforms others, achieving the \textbf{lowest} FVD of \textbf{173.463} and a competitive FID of \textbf{21.352}, indicating high quality and temporal coherence in animated talking heads. Additionally, it achieves the \textbf{highest} Sync-C score of \textbf{7.696} and the \textbf{lowest} Sync-D score of \textbf{7.536}, demonstrating excellent lip synchronization. These results highlight Teller’s effectiveness in maintaining both visual fidelity and synchronization.

\noindent \textbf{Comparison on the RAVDESS Dataset.} Table~\ref{tab:RAVDESS} shows quantitative results on the RAVDESS dataset. Our method again leads in performance, with the \textbf{lowest} FVD of \textbf{429.288} and a competitive FID of \textbf{20.352}, reflecting high-quality and temporally coherent animations. The \textbf{highest} Sync-C score of \textbf{4.496} and a competitive Sync-D score of \textbf{7.936} further demonstrate superior lip synchronization. These findings confirm Teller’s strength in producing synchronized and high-fidelity animated portraits.

\begin{figure}[t!]
    \centering
    \includegraphics[width=\linewidth]{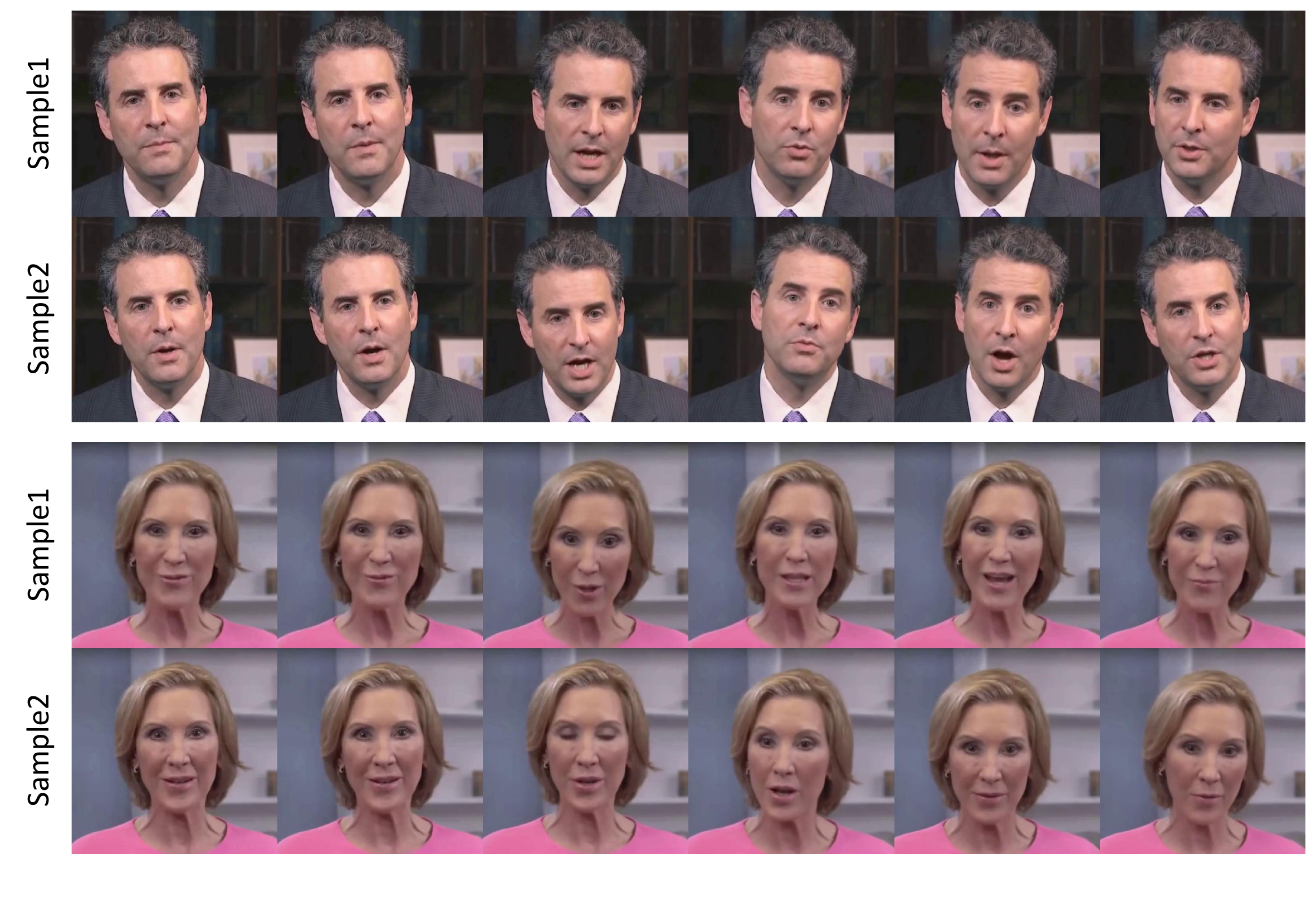}
    \vspace{-26pt}
    \caption{Top-k selection (k=15) in FMLG produces diverse facial expressions and actions with accurate lip sync on the HDTF.}
    \vspace{-12pt}
    \label{fig:sample_diversity}
\end{figure}

\begin{figure}[t!]
    \centering
    \includegraphics[width=\linewidth]{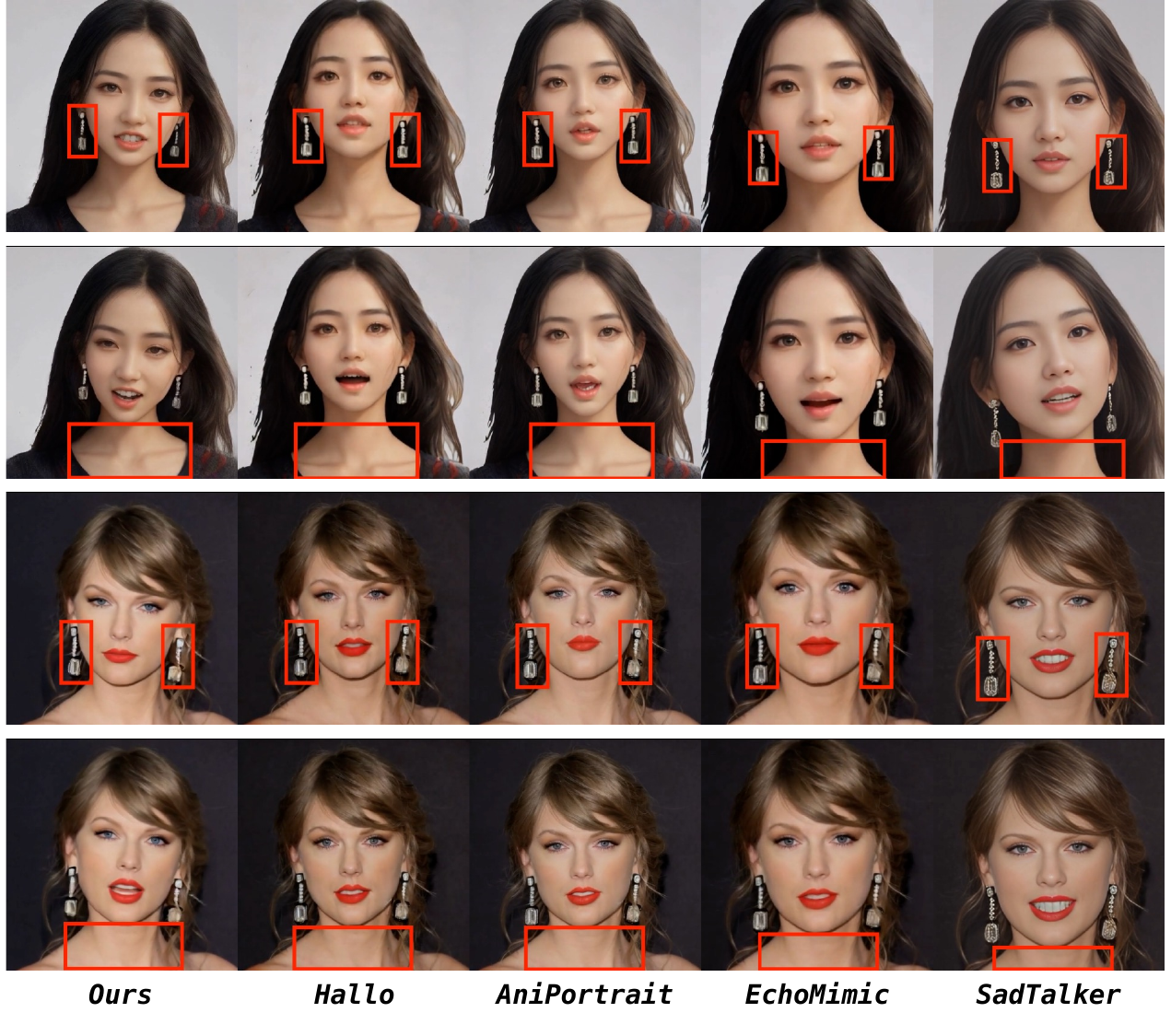}
    \vspace{-20pt}
    \caption{Visualization of finer motion details. \textit{\textbf{\textcolor{red}{Videos} are available in the supplementary materials.}}}
    \label{fig:movements}
    \vspace{-8pt}
\end{figure}

\begin{figure}[t!]
    \centering
    \includegraphics[width=\linewidth]{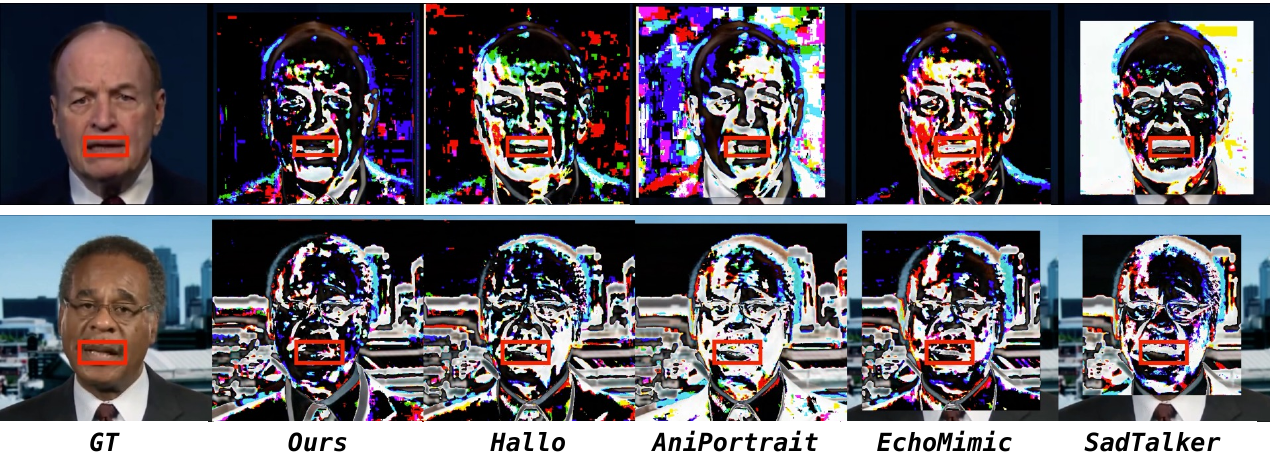}
    \vspace{-20pt}
    \caption{Visualization of the generation accuracy of lip shape.}
    \vspace{-12pt}
    \label{fig:Lip_acc}
\end{figure}

\begin{figure}[t!]
    \centering
    \includegraphics[width=\linewidth]{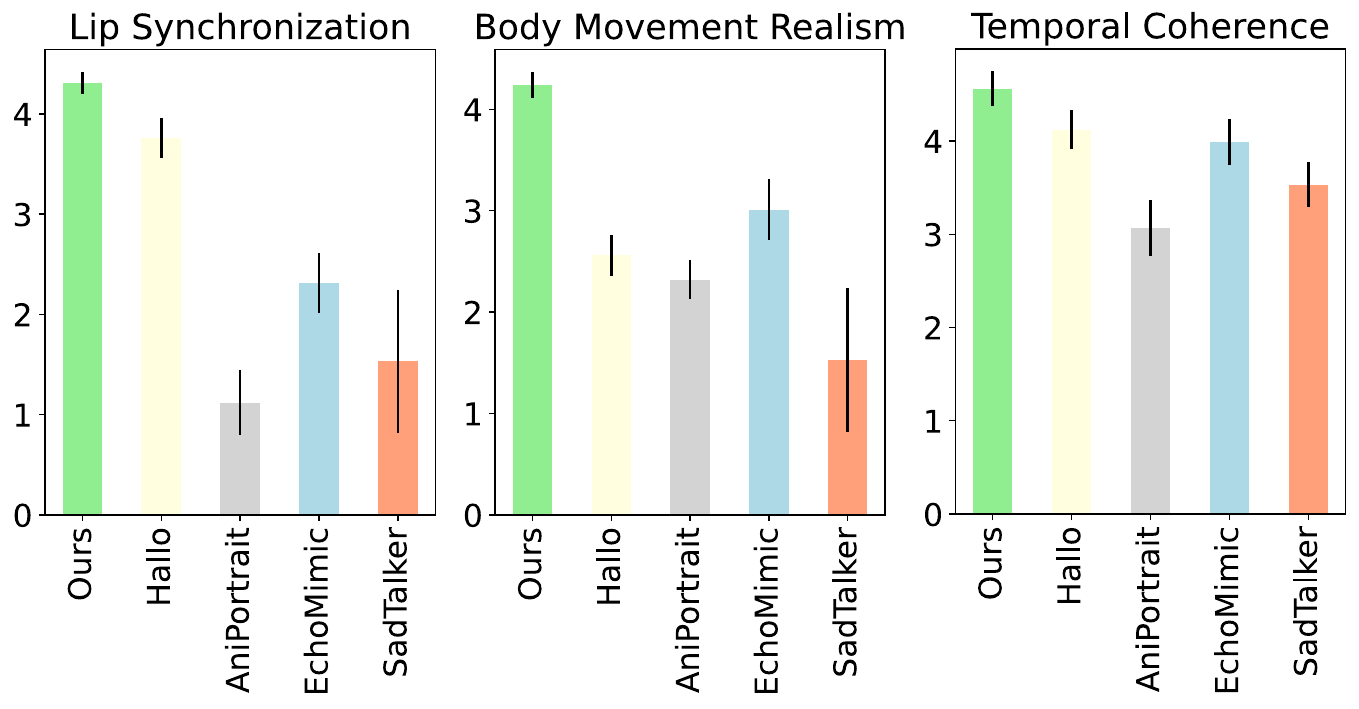}
    \vspace{-22pt}
    \caption{Human evaluation results among our proposed Teller and other SoTA methods.}
    \vspace{-20pt}
    \label{fig:human_eval}
\end{figure}
\subsection{Qualitative Results}
\noindent \textbf{Head Movement Comparison.}
Figure~\ref{fig:Qualitative_HDTF_Head}  shows a qualitative comparison of head movements. Teller replicates natural head movements more accurately, closely matching the ground truth (GT) with smooth, realistic turns and subtle expression-based adjustments. Competing methods, like AniPortrait and EchoMimic, often show abrupt or limited movements, appearing rigid or lifeless. Teller’s autoregressive framework ensures continuity and natural dynamics in head and emotion alignment, essential for lifelike animation. \\
\noindent \textbf{Diversity.} Fig.~\ref{fig:sample_diversity} shows two sequences generated with Top-k sampling (k=15) in FMLG, where each row represents frames from the same speech input. The model demonstrates diverse facial expressions and head movements while maintaining accurate lip sync, highlighting Top-k sampling’s role in enhancing motion variety without compromising synchronization. \\
\noindent \textbf{Emotional Expression.} Fig.~\ref{fig:Qualitative_RAVDESS_angry} presents our model has more accurate emotional expression ability due to the better speech understanding ability of AR transformer. \\
\noindent \textbf{Finer Motion Details.} Fig.~\ref{fig:movements} highlights fine motion details in neck and earring movements. Compared to other methods, Teller produces realistic, nuanced motions synchronized with speech, capturing subtle audio-driven dynamics that contribute to lifelike and temporally coherent animation. The consistent detail across frames underscores Teller’s robustness. \\
\noindent \textbf{Lip Synchronization Accuracy.} Fig.~\ref{fig:Lip_acc} shows superior lip synchronization, with Teller aligning generated lip shapes closely to natural movements. This high fidelity highlights Teller’s strength in producing realistic, synchronized mouth motions for convincing talking head animation.






\subsection{Human Evaluation}
We conducted a human evaluation to assess the quality of generated animations, focusing on lip synchronization, body movement realism, and temporal coherence. Thirty participants (66.7\% aged 24-30, 33.3\% aged 30-40; 30\% male, 70\% female; 83.3\% with AIGC model experience) rated each animation on a 5-point Likert scale for coherence with input and animation quality. A total of 100 videos were presented in random order to avoid bias, providing insights into subjective perceptions of animation quality and natural expression alignment. As shown in Figure~\ref{fig:human_eval}, participants rated Teller highest in lip synchronization, body movement realism, and temporal coherence, with low variance in scores, indicating robust and consistent performance.

%% file: sec/5_ab.tex
\section{Ablation Study}
\begin{figure}[t!]
    \centering
    \includegraphics[width=\linewidth]{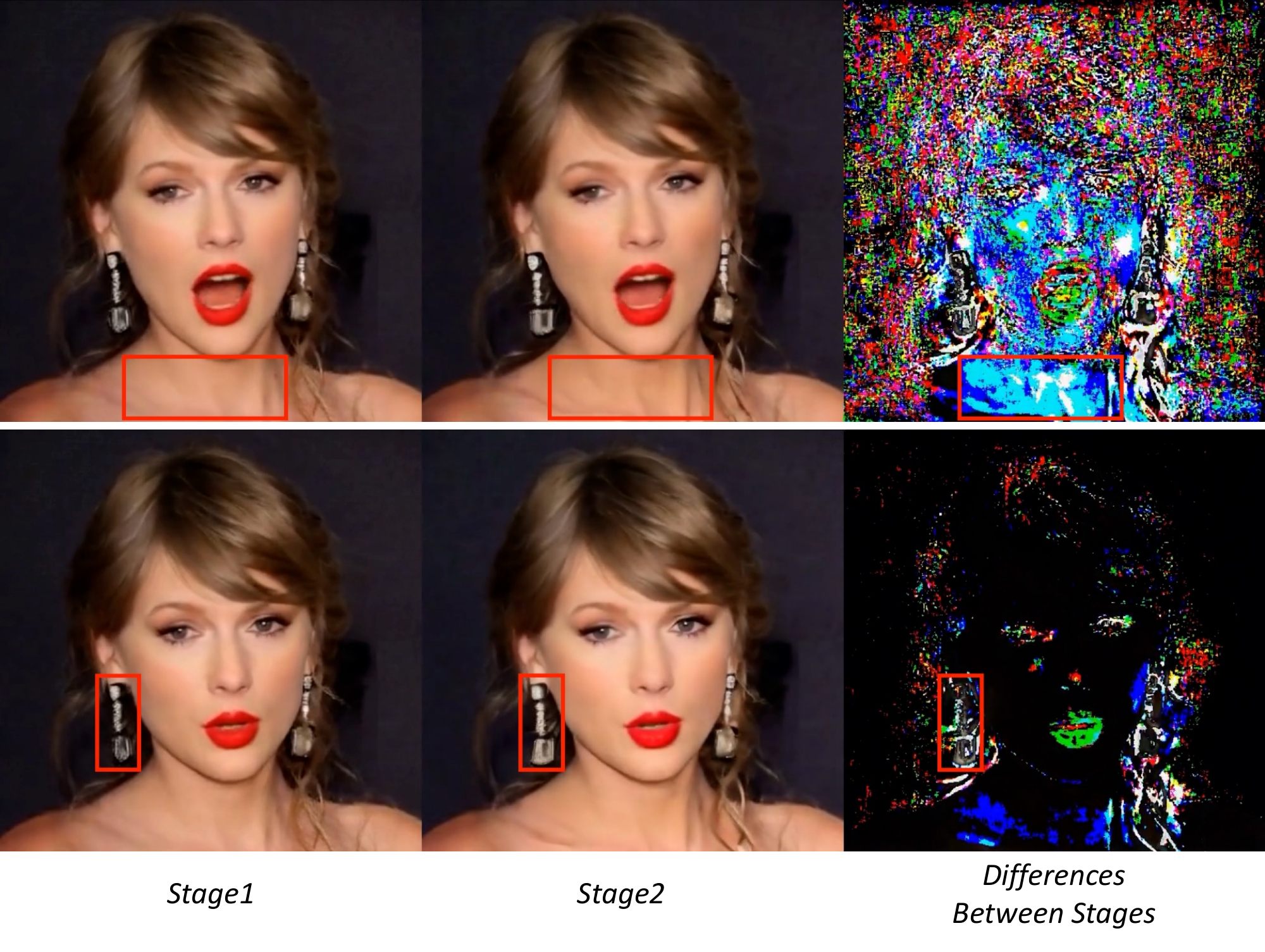}
    \vspace{-24pt}
    \caption{Visualization of the generation images with different stages and the differences between stages.}
    \vspace{-20pt}
    \label{fig:module_ablation}
\end{figure}

\noindent \textbf{Stage (Module) Ablation.} We perform an ablation study comparing Stage 1 and Stage 2, focusing on animation quality in body parts and accessories, especially neck muscles and earrings. As shown in Fig.~\ref{fig:module_ablation}, Stage 1 outputs often show less realistic, inconsistent movements. In contrast, Stage 2, with ETM, significantly improves the physical consistency of subtle motions, creating natural earring sway and smooth neck movements for lifelike, temporally coherent animation.

\begin{figure}[t!]
    \centering
    \vspace{+8pt}
    \includegraphics[width=\linewidth]{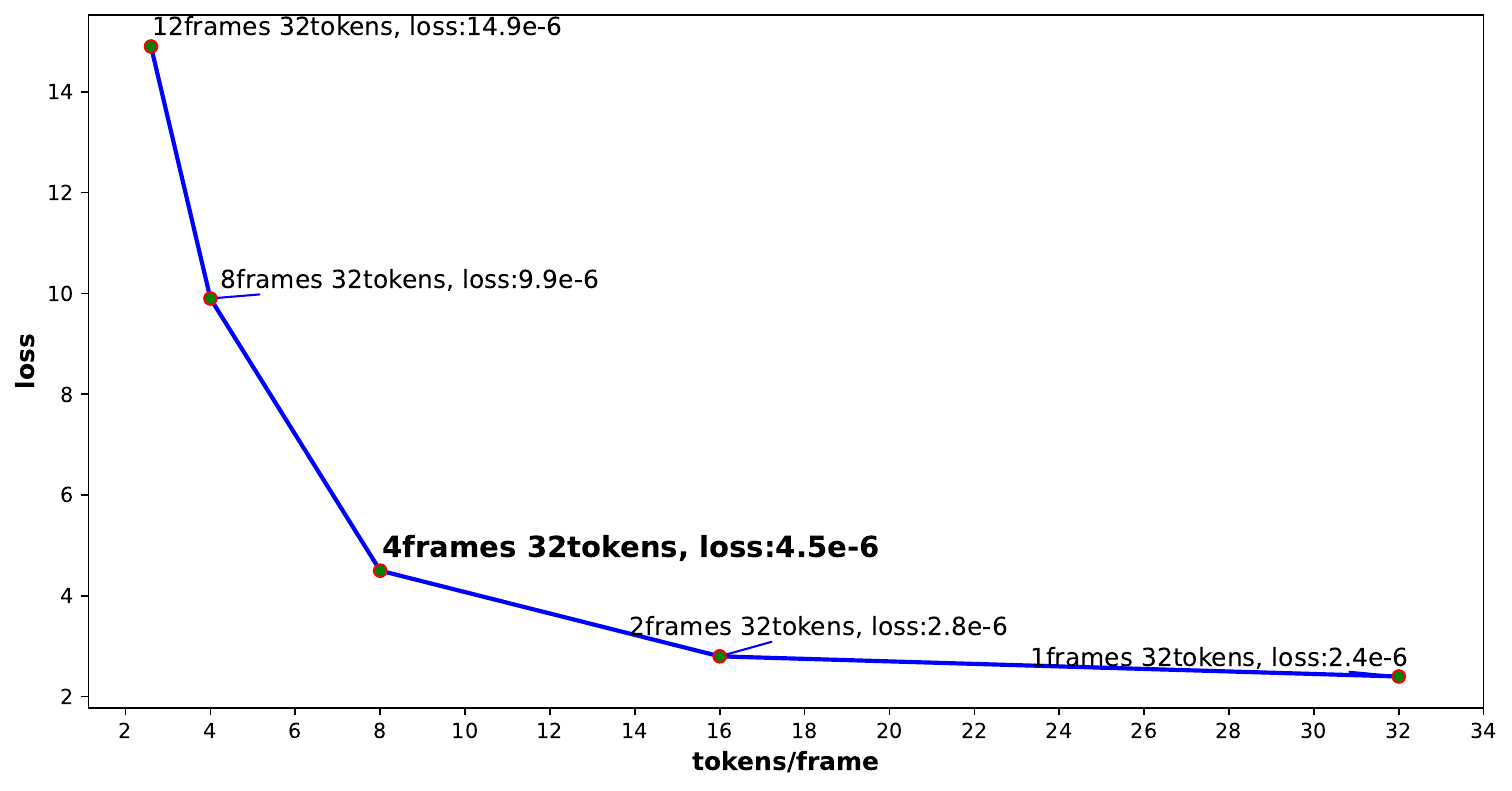}
    \vspace{-19pt}
    \caption{Tradeoff between performance (loss) and different compression(tokens/frame ) ratios.}
    \vspace{-20pt}
    \label{fig:VQ_ablation}
\end{figure}

\noindent \textbf{Frames / Tokens in  RVQ  Module}
Figure~\ref{fig:VQ_ablation} shows the impact of frame and token configurations in the  RVQ  module on face shape accuracy in talking head animations. This ablation study explores how varying frames and tokens affects facial movement quality and synchronization. Results show that increasing frames and tokens improves facial dynamics, enhancing lip synchronization and realism, though with higher computational costs. We selected 4 frames and 32 tokens for an optimal balance, adjustable as needed.

\noindent \textbf{Ablation Study on Audio Condition Encoder.} We analyzed the performance differences between TTS and ASR models as audio condition encoders for our task. Specifically, we used Whisper for the ASR model and funcodec for the TTS model. As shown in Table~\ref{tab:audio_encoder_comparison}, Whisper achieved a Sync-C score of 7.696 and a Sync-D score of 7.536, while funcodec scored 4.286 in Sync-C and 10.373 in Sync-D. These results indicate that our task benefits more from the ASR model's capability to capture the nuances required for precise synchronization, as seen in Whisper’s higher Sync-C and lower Sync-D scores. This analysis suggests that the ASR model is better suited as the audio condition encoder, enhancing the overall quality and synchronization of our talking head animation.

\begin{table}[t!]
\centering
\vspace{-8pt}
\begin{tabular}{lcc}
\toprule
Method & Sync-C $\uparrow$ & Sync-D $\downarrow$ \\
\midrule
w funcodec & 4.286& 10.373\\
w whisper  & \textbf{7.696}& \textbf{7.536}\\
\bottomrule
\end{tabular}
\vspace{-8pt}
\caption{Comparison of synchronization for audio conditions using funcodec and Whisper in ASR and TTS tasks on HDTF.}
\label{tab:audio_encoder_comparison}
\end{table}

\noindent \textbf{Ablation Study on Single-Head vs. Multi-Head Architecture.}  
We compare single-head and multi-head models on FID, FVD, Sync-C, and Sync-D metrics (Table~\ref{tab:onehead_mulhead_comparison}). The single-head model slightly outperforms in FID (22.110 vs. 21.352) and has comparable FVD (172.553 vs. 173.463), with marginally better synchronization (Sync-C of 7.790 vs. 7.696 and Sync-D of 7.474 vs. 7.536). Both architectures yield competitive results, though the single-head model slightly excels in synchronization, while the multi-head model offers greater real-time potential .

\begin{table}[t!]
\centering
\vspace{-8pt}
\begin{tabular}{lllcc}
\toprule
Method &   \textbf{FID$\downarrow$}&\textbf{FVD$\downarrow$}&\textbf{Sync-C$\uparrow$}& \textbf{Sync-D$\downarrow$}\\
\midrule
Single-Head&   22.110&\textbf{172.553}&\textbf{7.790}& \textbf{7.474}\\
Multi-Head&   \textbf{21.352}&173.463&7.696& 7.536\\
\bottomrule
\end{tabular}
\vspace{-10pt}
\caption{Comparison of  performance between Single-Head and Multi-Head on the HDTF dataset.}
\vspace{-12pt}
\label{tab:onehead_mulhead_comparison}
\end{table}